\documentclass[letterpaper, 10 pt, conference]{ieeeconf}  

\IEEEoverridecommandlockouts                              

\usepackage{graphics} 
\usepackage{epsfig} 
\usepackage{amsmath} 
\usepackage{amssymb}  
\usepackage{cite}
\usepackage{graphicx, subfig}
\usepackage[ruled,vlined,linesnumbered]{algorithm2e}
\usepackage{multirow}
\usepackage{dcolumn}
\newcolumntype{d}[1]{D{.}{.}{#1}}
\usepackage{longtable}
\usepackage{makecell}
\usepackage{threeparttable}
\usepackage{diagbox}
\usepackage{float}
\usepackage{booktabs}
\usepackage{stfloats}

\title{\LARGE \bf
	DetFlowTrack: 3D Multi-object Tracking based on Simultaneous Optimization of Object Detection and Scene Flow Estimation
}

\author{Yueling Shen, Guangming Wang, and Hesheng Wang 
\thanks{*This work was supported in part by the Natural Science Foundation of China under Grant 62073222 and U1913204, in part by “Shu Guang”project supported by Shanghai Municipal Education Commission and Shanghai Education Development Foundation under Grant 19SG08,  in part by Shenzhen Science and Technology Program under Grant JSGG20201103094400002, in part by the Science and Technology Commission of Shanghai Municipality under Grant 21511101900,  in part by grants from NVIDIA Corporation. Corresponding Author: Hesheng Wang.}
\thanks{The authors are with Department of Automation, Key Laboratory of System Control and Information Processing of Ministry of Education, Key Laboratory of Marine Intelligent Equipment and System of Ministry of Education, Shanghai Engineering Research Center of Intelligent Control and Management, Shanghai Jiao Tong University, Shanghai 200240, China.}%
}

\begin{document}

\maketitle
\thispagestyle{empty}
\pagestyle{empty}

\begin{abstract}
3D Multi-Object Tracking (MOT) is an important part of the unmanned vehicle perception module. Most methods optimize object detection and data association independently. These methods make the network structure complicated and limit the improvement of MOT accuracy. 
we proposed a 3D MOT framework based on simultaneous optimization of object detection and scene flow estimation. In the framework, a detection-guidance scene flow module is proposed to relieve the problem of incorrect inter-frame assocation. For more accurate scene flow label especially in the case of motion with rotation, a box-transformation-based scene flow ground truth calculation method is proposed. 
Experimental results on the KITTI
MOT dataset show competitive results over the state-of-the-arts and the robustness under extreme motion with rotation.
\end{abstract}

\section{INTRODUCTION}
3D MOT is an important perception technique required for the applications of autonomous driving\cite{chiu2020probabilistic, 9341164}. 3D MOT takes continuous frame sequence as input and outputs trajectories of targets in specific categories. The trajectory of a target is represented by 3D boxes with tracking id in consecutive frames. The same target is identified with the corresponding unique ID in different frames. 

Most methods\cite{9341164, shenoi2020jrmot, chiu2020probabilistic, zhai2020flowmot, 9157602} follow the tracking-by-detection framework, which split the task into two stages: object detection and data association. With the object detection results, the data association module calculates the similarities between each pair in adjacent frames, and the same target is identified with the same unique ID. The tracking-by-detection framework is optimized independently for detection and data association. It limits the improvement of tracking accuracy and makes the whole process redundant and complex. 

Traditional data association methods\cite{9341164, shenoi2020jrmot, chiu2020probabilistic} associate boxes in adjacent frames with motion prediction methods such as Kalman filters or particle filters. Those filter-based methods are vulnerable in extreme motion conditions such as sudden braking or turning. What is more, the fixed motion parameters in these methods are not suitable for multi-categories MOT because different types of targets may have different motion models. 

Optical flow\cite{7410673, 8579029} and Scene flow\cite{wu2019pointpwc, 8953876} estimates the inter-frame motion of points. Inter-motion of objects instead of points are estimated in MOT. Considering the correlation between scene flow estimation and MOT, many studies\cite{zhang2020multiple, zhai2020flowmot, 9000527} estimate the inter-frame motion of objects with scene flow estimation.
FlowMOT\cite{zhai2020flowmot} directly replaces the motion estimation model with the trained scene flow model, improves the robustness under extreme motion. However, it is still optimized independently for detection and scene flow estimation. PointTrackNet\cite{9000527} proposed an end-to-end 3D object detection and tracking network. This method optimizes object detection firstly, followed by data association based on scene flow estimation. It does not consider the problem of incorrect association between different targets in data association. The scene flow label is not accurate enough in the case of rotation of the target. All these problems limit the accuracy of MOT results. 

In order to solve the above problems, we propose a 3D multi-object tracking framework based on simultaneous optimization of object detection and scene flow estimation to simplify the network structure, improve the MOT accuracy and improve the robustness under extreme motion with rotation. The contributions of our work are listed as follows:
\begin{itemize}
	\item A new framework of 3D MOT that simultaneously optimizes object detection and scene flow estimation  is proposed. It takes two adjacent point clouds as input and outputs MOT trajectories.
	\item A detection-guidance scene flow module is proposed to relieve incorrect inter-frame association of points belonging to different objects.
	\item A box-transformation based scene flow label calculation method is proposed to improve the robustness in the presence of extreme motion with rotation.
\end{itemize}

\section{THE PROPSED METHOD}
\subsection{Overall Framework}
Instead of independent optimization of object detection and scene flow estimation, we proposed a 3d multi-object tracking framework that simultaneously optimizes object detection and scene flow estimation. 
The proposed framework takes point cloud sequences as inputs, and outputs objects' trajectories. 

\begin{figure*}[t]  
	\centering
	\includegraphics[width=0.9\textwidth]{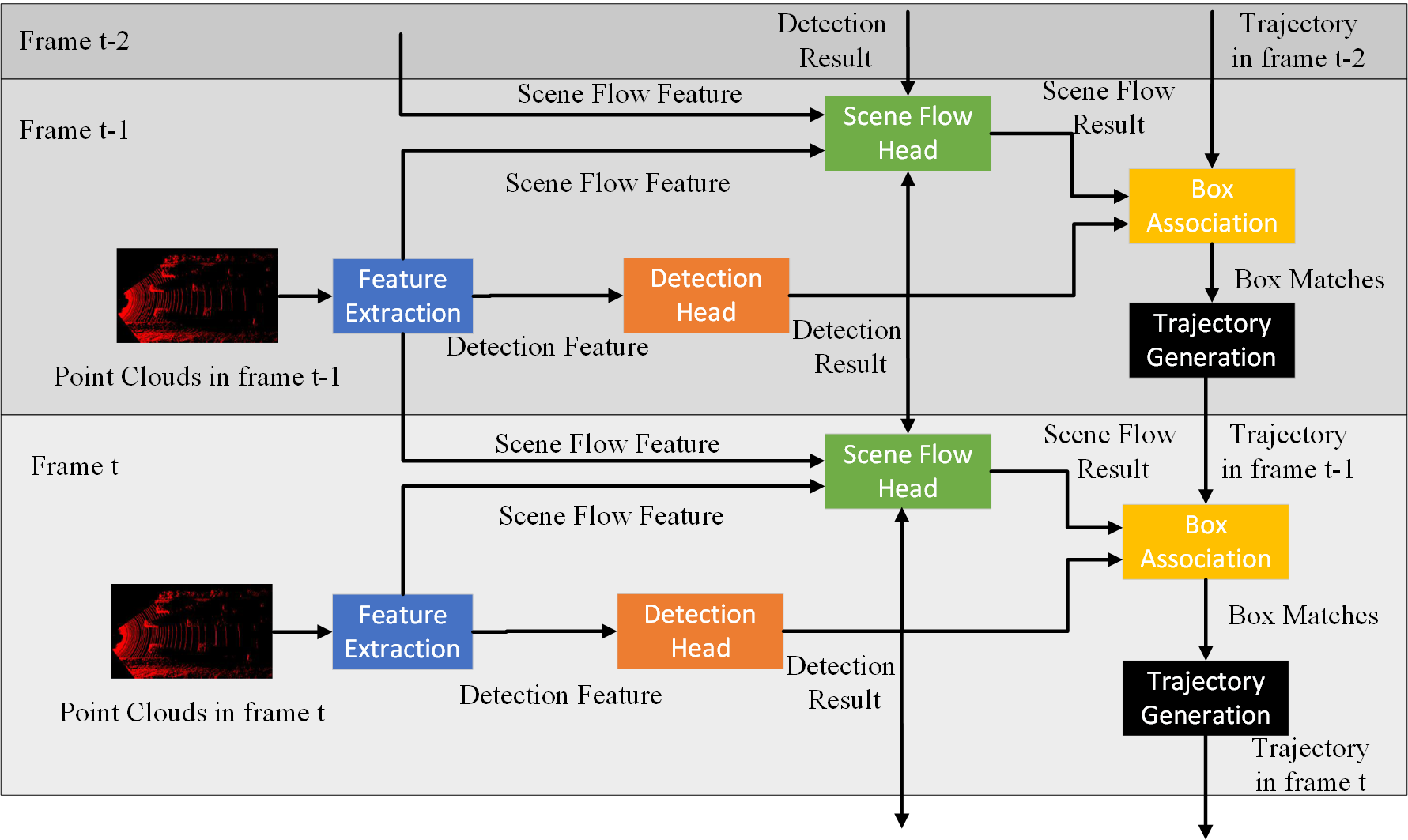} 
	\caption{Overall Framework. With point cloud as input, point-wise detection feature and scene flow feature are extracted simultaneously by feature extraction module. In the guidance of detection result, the scene flow feature in two adjacent frames is input into the scene flow head to estimate the inter-frame point motion. With the inter-frame points' motion estimation, inter-frame boxes' motion is estimated in box association. The inter-frame boxes motion is used to make inter-frame boxes associated and finally generate the trajectory.} 
	\label{inference_framework} 
\end{figure*}

As shown in figure\ref{inference_framework}, point-wise data association between adjacent frames are achieved by scene flow head to avoid dependence on motion model parameters.
To simplify the network structure, we extract both object detection feature and scene flow feature in the feature extraction module. During the training stage, the loss of object detection and scene flow will be weighted summed. The total loss will be used to optimize the overall network. 
Considering that the detection results contain semantic information of points, we input the detection results into the scene flow module as a guide to avoid wrong data association of points belonging to different objects.
To achieve multi-object tracking, boxes between frames are matched according to the result of object detection and scene flow in the box association module. The trajectory generation module takes box matches as input and updates trajectory.

\subsection{Fearute Extraction}
\begin{figure}[t]
	\begin{center}
		\includegraphics[width=0.47\textwidth]{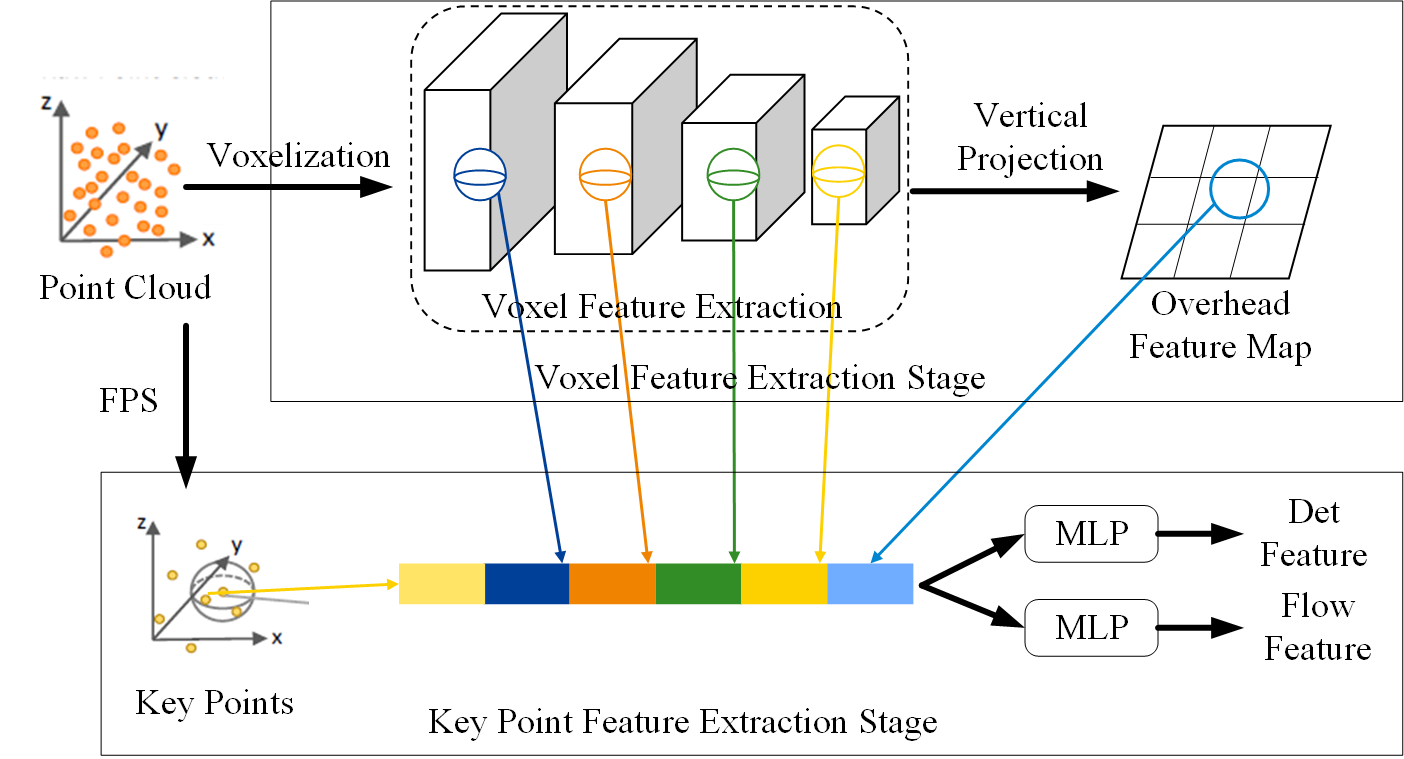}
	\end{center}
	\caption{The Framework of Feature Extraction}
	\label{featue_extraction}
\end{figure}

To extract the point-wise detection feature and flow feature of the point cloud simultaneously, we utilize the backbone of PV-RCNN\cite{9157234}. As shown in figure \ref{featue_extraction}, voxel feature at multiple resolutions is extracted by voxel feature extraction. After that, the overhead feature map is obtained from the last layer of the feature voxel by vertical projection. We downsample the point clouds by farthest point sampling and key points are extracted. In the key point feature extraction stage, multiple resolutions voxel features are summarized into the feature embeddings of a small set of key points. Two parallel multi-layer perceptions take the key point feature embeddings as input and output detection feature and scene flow feature simultaneously.

\subsection{Detection Head}
For accurate detection results, we apply a two-stage detection head as PV-RCNN\cite{9157234}. In the first stage, 3D box proposals are generated from the overhead feature map. In the proposal refine stage, the 3D box proposals are refined by detection features of points inside. For 3D box proposal $i$, the outputs of detection head include box result $\emph{b}_i=(px_i, py_i, pz_i, w_i, l_i, h_i, \theta_i, c_i)$ and score result $\emph{s}_i$.
\subsection{Scene Flow Head}
The scene flow head contains the cost layer and scene flow calculation layer. The cost layer realizes feature aggregation between two point clouds, and outputs cost. The scene flow calculation layer takes cost volume as input, outputs scene flow result for each point in frame t-1. 


\subsubsection{Cost Layer}
The inputs of the cost layer are two adjacent key point clouds: $P^{t-1}=\{\emph{p}_i^{t-1}=(x_i^{t-1},y_i^{t-1},z_i^{t-1})\}_{i=1}^{N}$ and $P^{t}=\{\emph{p}_i^{t}=(x_i^{t},y_i^{t},z_i^{t})\}_{i=1}^{N}$, the corresponding flow feature: $F_{flow}^{t-1}=\{\emph{f}_i^{t-1}\}_{i=1}^{N}; F_{flow}^{t}=\{\emph{f}_i^{t}\}_{i=1}^{N}$ and the corresponding detection result: $O_{det}^{t-1}=\{\emph{o}_i^{t-1}\}_{i=1}^{N}; O_{det}^{t}=\{\emph{o}_i^{t}\}_{i=1}^{N}$, where $N$ is the number of key points in each frame and t ranges from 2 to the number of frames. The flow feature vector $\emph{f}_i^{t} \in \mathbb{R}^{C_{flow}}$ and the detection result vector $\emph{o}_i^{t}$ is achieved by concatenating box vector $\emph{b}_i^{t}$ and score ${s}_i^{t}$ in detection result. The cost layer outputs cost volume $\emph{c}_i^{t}$ associated with point $\emph{p}_i^{t-1}$ in frame t-1. 
\begin{figure}[t]
	\begin{center}
		\includegraphics[width=0.5\textwidth]{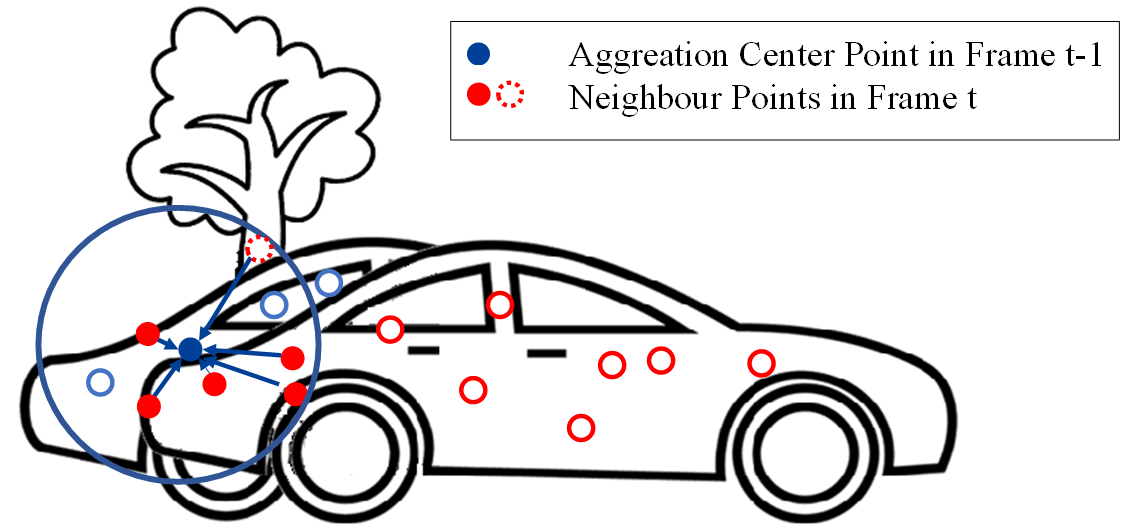}
	\end{center}
	\caption{The Visualization of Cost Layer}
	\label{cost_layer}
\end{figure}

Figure \ref{cost_layer} is the visualization of the cost layer. For point $\emph{p}_i^{t-1}$, we find $K$ nearest points in frame $t$, denoted as $N_i^{t}=\{\emph{p}_j^{t}\}_{j=1}^{K} \subset P^{t}$. 

For point $\emph{p}_i^{t-1}$ as the center point, feature of neighbor points $N_i^{t}$ are aggregated to learn the relative motion between frames. 

The previous work\cite{9000527} concates the direction vector $(\emph{p}_i^{t-1}-\emph{p}_j^t)$, the flow feature of neighbour point $\emph{f}_j^{t}$ and the flow feature of center point $\emph{f}_i^{t-1}$ as embedded features, and then applies multi-layer perceptions and element-wise max pooling to directly achieve the point-wise association tracking displacement. 

This method takes all neighbor points into consideration during aggregation. For accurate motion estimation, only the neighbor points belonging to the same object with point $\emph{p}_i^{t-1}$ should be included. As shown in figure \ref{cost_layer}, the red point with dotted edges belongs to a tree, while the center point belongs to a car. Although within neighbor range, aggregating the point belong to different object may disturb the relative motion estimation.

Previous scene flow estiamtion network\cite{wu2019pointpwc} takes the embedded features as input, the cost between point $\emph{p}_i^{t-1}$ and its neighbour point $\emph{p}_j^{t}$ can be defined as equation \ref{cost_ij}, where $h(\cdot)$ is a function with learnable parameters. 
\begin{equation}
	\operatorname{Cost}\left(i, j\right)=h\left(\emph{f}_i^{t-1}, \emph{f}_j^{t}, \emph{p}_i^{t-1}-\emph{p}_j^t\right).
	\label{cost_ij}
\end{equation}

Feature aggregation of point $\emph{p}_i^{t-1}$ is conducted by weighted sum the cost of all the neighbour points as defined in equation \ref{cost_i_t-1}, where $w_(i,j)$ denotes the weight for match: $(\emph{p}_i^{t-1}, \emph{p}_j^{t})$. After feature aggregation, we obtain cost volume $\emph{c}_i^{t-1}$ for point $\emph{p}_i^{t-1}$.
\begin{equation}
	\emph{c}_i^{t-1}=\sum_{\emph{p}_j^{t} \in N_{i}^{t}} w(i,j) \cdot  h\left(\emph{f}_i^{t-1}, \emph{f}_j^{t}, \emph{p}_i^{t-1}-\emph{p}_j^t\right).
	\label{cost_i_t-1}
\end{equation}

The weight $w(i,j)$ is calculated as shown in equation \ref{point_pwc_weight}. It is learned as a continuous function of the directional vectors $(\emph{p}_i^{t-1}-\emph{p}_j^t)$ with multi-layer perceptions. 
\begin{equation}
	w(i,j)=M L P\left(\emph{p}_i^{t-1}-\emph{p}_j^t\right).
	\label{point_pwc_weight}
\end{equation}

Those weight calculation method only takes directional vectors as inputs. As a result, the closer the neighbor point is to the center, the greater the weight.

Considering that the detection result contains the object information the point belongs to, we use the guidance of detection results in feature aggregation. 
Our weight is calculated by equation \ref{w_e}-\ref{w}, where $w_g(i,j)$ denotes the reciprocal of distance in geometric space, $w_o(i,j)$ denotes the reciprocal of distance in detection result space, $MEAN(x,y)$ denotes the average of $x$ and $y$, and the weight $w(i,j)$ is calculated by the average of normalized $w_g(i,j)$ and normalized $w_o(i,j)$. As a result, only nearby points belonging to the same object are given greater weight.

\begin{equation}
	w_e(i,j)=\frac{1}{\|{\emph{p}_i^{t-1}-\emph{p}_j^{t}}\|_{2}}
	\label{w_e}
\end{equation}
\begin{equation}
	w_o(i,j)=\frac{1}{\|{\emph{o}_i^{t-1}-\emph{o}_j^{t}}\|_{2}}
	\label{w_o}
\end{equation}
\begin{equation}
	w(i,j)=MEAN(\frac{w_e(i,j)}{\sum_{j=1}^{K} w_e(i,j)}, \frac{w_o(i,j)}{\sum_{j=1}^{K} w_o(i,j)})
	\label{w}
\end{equation}

Substituting equation \ref{w} to equation \ref{cost_i_t-1}, the cost volume $c_i^{t-1}$ contains the inter-frame motion information of point $\emph{p}_i^{t-1}$.

\subsubsection{Scene Flow Calculation Layer}
The input of the scene flow calculation layer are cost volume set $C^{t-1}=\{\emph{c}_i^{t-1}\}_{i=1}^{N}$, the flow feature set of the first point cloud $F_{flow}^{t-1}$, and the first point cloud $P^{t-1}$. 
Firstly, point-wise concatenating of cost volume and flow feature is conducted. After that, the PointConv\cite{8954200} layers are applied to merge the flow feature locally, following MLP layers to estimate scene flow $D^{t-1,t}$ corresponding to the first point cloud $P^{t-1}$.
\subsubsection{Scene Flow Label Calculation}
In the practical application, the ground truth label of scene flow is hard to obtain. Aiming to solve this problem, we propose an MOT-induced scene flow label calculation algorithm. MOT ground truth label provides detection results with tracking id for objects in each frame. The tracking id of the same object keeps the same during sequences. As a result, the detection results in two adjacent frames can be matched according to the tracking id, so as to calculate the inter-frame relative pose transformation of the object. Under the assumption of rigid-body transformation, the pose transformation of the point belonging to the object is equal to the pose transformation of the object. 

PointTrackNet\cite{9000527} calculates the scene flow label by substituting the bounding boxes’ movement between two frames. This box-translation based method is not accurate enough when the target rotates between frames. We propose a box-tranformation based method to obtain more accurate scene flow ground truth label. For point $\emph{p}_i^{t-1}$ belonging to box $m$, we transform it with the box transformation $T_m$ and obtain the point $\tilde{\emph{p}}_i^{t-1}$. We obtain the scene flow ground truth label $\hat{d}^{t-1}_i$ as eqaution \ref{d_gt}.
\begin{equation}
	\hat{d}^{t-1}_i=\tilde{\emph{p}}_i^{t-1}-\emph{p}_i^{t-1}
	\label{d_gt}
\end{equation}

\subsection{Box Association}
Box association realizes inter-frame motion estimation of objects with the inter-frame motion estimation of points from scene flow results. 
Firstly, the inter-frame transformation of trajectory each box is predicted by its internal points' scene flow results. And then, the predicitive trajectory box is obtained by transforming the previous frame trajectory box with predicted transformation. The matching cost matrix between the predictive trajectory boxes and the detection boxes is calculated by the boxes's intersection of union. Finally, the Hungarian algorithm\cite{kuhn1955hungarian} is applied with the matching cost matrix and outputs match between the predictive trajectory boxes and the detection boxes. 

The key to the box association is to achieve inter-frame boxes' motion estimation according to inter-frame points' motion results from scene flow. The process of box association is the reverse of the the process of scene flow groundtruth label calculation. Corresponding to the scene flow label calculation, we solve the boxes' inter-frame transformation based on point cloud registration. Specifically, for each trajectory box $b_i^{t-1}$, its internal points are extracted and form in-box point cloud $c_i^{t-1}$. For each point in $c_i^{t-1}$, translate it with its scene flow and form the predictive in-box point cloud $\tilde{c}_i^{t-1}$. Then, the transformation $T_i^{t-1,t}$ between $c_i^{t-1}$ and $\tilde{c}_i^{t-1}$ is calculated based on svd method. The transformation $T_i^{t-1,t}$ is applied to trajectory box $b_i^{t-1}$ and obtain the predictive trajectory box $\tilde{b}_i^{t-1}$.
\subsection{Trajectory Generation}
%
%

The matching result from box association module is input into the trajectory generation module to update the trajectory. The matching result is represented by three lists: match list $m^{t-1,t}$, detection box unmatch list $um^{t-1,t}_{(d)}$ and trajectory box unmatch list $um2^{t-1,t}_{(t)}$. The match list $m^{t-1,t}$ consists of matching pairs of detection boxes and trajectory boxes. The unmatch list $um^{t-1,t}_{(d)}$ and $um^{t-1,t}_{(t)}$ consists of unmatch boxes of detection boxes and presictive trajectory boxes separately. In order to cope with occlusion and error detection, track boxes are added and deleted according to the times of matching and losing. 

\subsection{Loss Function}
The loss function consists of two parts: detection loss and scene flow loss as shown in equation \ref{loss}. 
\begin{equation}
	{L}=L_{(det)}+L_{sceneflow}.
	\label{loss}
\end{equation}

In PV-RCNN\cite{9157234}, the detection loss can be split into the region proposal loss $L_{rpn}$, the keypoint segmentation loss $L_{seg}$ and the proposal refinement loss $L_{rcnn}$. In our work, we fix the parameters of RPN. As a result, our detection loss is
calculated as equation \ref{loss_det}.
\begin{equation}
	{L_{det}}=L_{seg}+L_{rcnn}.
	\label{loss_det}
\end{equation}

For scene flow loss calculation, we calculate in-box mask $M^{(\mathrm{inbox})}_{i}$ for each point according to the MOT ground truth label. Scene flow loss is calculated by root mean square error (RMSE) as equation \ref{loss_sceneflow}. Only points belonging to objects are considered. 
\begin{equation}
	L^{(\text {sceneflow })}=\sqrt{\frac{\sum_{i=1}^{n} M_{i}^{(\mathrm{inbox})} \times\left(d_{i}-\hat{d}_{i}\right)^{2}}{N_{(\mathrm{inbox})}}},
	\label{loss_sceneflow}
\end{equation}
where $d_{i}$ and $\hat{d}_{i}$ are scene flow estiamtion result and scene flow label for point i. $N_{(\mathrm{inbox})}$ is the number of points inside object boxes.

\section{EXPERIMENTS RESULTS AND DISCUSSIONS}
\subsection{Dataset and Evaluation Metirces}
We train and evaluate our DetFlowTrack on KITTI MOT dataset\cite{6248074}. KITTI MOT dataset consists of 21 training sequences and 29 testing sequences. Each sequence provides a sequence of point clouds collected by lidar and a sequence of images collected by color monocular cameras. We split the training sequences into two parts: Seq-0000 to Seq-0015 for training and Seq-0016 to Seq-0020 for validation. 

To improve sample diversity and avoid overfitting, we intorduced a data augmentation approach for both object detection and scene flow estimation. First, we generate a object database with the KITTI Object dataset\cite{6248074} by intercepting the labels and the in-box point cloud of all the groundtruth targets. Then during training, several groundtruth objects are randomly selected from the object database and are put on to the same random location of two adjacent frames' scene. Using this approach, the number of objects and the diversity of the objects' location is greatly increased. After that, global flip, rotation with random angle and  scaling with random ratio. To increase the diversity of objects' motion between frames, random traslation is applied to each objects independently.

We adopt HOTA\cite{luiten2021hota} (Higher Order Tracking Accuracy) as the evaluation metric. The HOTA metric can split into DetA (Detection Accuracy Acore), AssA (Association Accuracy Score) and LocA (Localization Accuracy Score). 
\subsection{Network Details}
We fix the parameters of RPN in PV-RCNN backbone and train detection network and scene flow estimation network at the same time during training. We use an Adam optimizer with one cycle learning rate\cite{smith2019super}. The learning rate changes a cosine function with 0.01 upper bound. 

We train on an Intel 3.40GHz Gold 6128 CPU and 4 GeForce
GTX 1080Ti GPU with batch size=4. Our DetFlowTrack is implemented on pytorch with CUDA 10.0.
\subsection{Abation Atudy}
An ablation study is designed to verify the validity of the proposed data augmentation module and the detection results' guidance in scene flow module. The results are show in table \ref{tab:ablation_study}. In the catagory of both car and pedestrian, the proposed data augmentation module and the detection results' guidance in scene flow module are effective to improve the accuracy of MOT. 
%

\begin{table}[t]
	\centering
	\caption{Ablation Study Results for Data Augmentation and Detection Results' Guidance}
	\begin{tabular}{ccc}
		\toprule
		Data Augmentation & Detection Results' Guidance & HOTA$\uparrow$ \\
		\midrule
		\multicolumn{3}{c}{Car Category} \\
		\midrule
		&       & 75.89\% \\
		$\checkmark$     &       & 77.43\% \\
		$\checkmark$     & $\checkmark$     & 78.80\% \\
		\midrule
		\multicolumn{3}{c}{Pedestrian Category} \\
		\midrule
		&       & 42.77\% \\
		$\checkmark$     &       & 46.42\% \\
		$\checkmark$     & $\checkmark$     & 78.80\% \\
		\bottomrule
	\end{tabular}%
	\label{tab:ablation_study}%
\end{table}%

\subsection{Quantitative Results}
We tested the trained model on KITTI MOT test dataset and submitted the test results to KITTI website for accuracy evaluation. The proposed algorithm is named FlowDetTrack and the results are compared with the latest online MOT algorithm. Table \ref{tab:test} shows the accuracy evaluation results. The HOTA of the proposed method in this paper far exceeds that of the state-of-the-art methods with point cloud input. 

%

\begin{table}[t]
	\centering
	\caption{Evaluation Results in Test Sets of KITTI MOT. The bold font highlight the best results.}
	\begin{tabular}{ccccc}
		\toprule
		\multicolumn{5}{c}{Car Category} \\
		\midrule
		Method & HOTA$\uparrow$ & DetA$\uparrow$ & AssA$\uparrow$ & LocA$\uparrow$ \\
		\midrule
		Complexer-YOLO\cite{simon2019complexer}  & 49.12\% & 62.44\% & 39.34\% & 81.47\% \\
		AB3DMOT\cite{9341164}  & 68.61\% & 71.06\% & 69.06\% & 86.80\% \\
		PointTrackNet\cite{9000527}  & 57.20\% & 55.71\% & 59.15\% & 80.07\% \\
		FlowDetTrack(ours) & \textbf{71.52\%} & \textbf{72.87\%} & \textbf{70.89\%} & \textbf{87.79\%} \\
		\midrule
		\multicolumn{5}{c}{Pedestrian Category} \\
		\midrule
		Complexer-YOLO\cite{simon2019complexer}  & 14.08\% & 24.91\% & 8.15 \%	 & 68.64\% \\
		AB3DMOT\cite{9341164}  & 35.57\% & 32.99\% & 38.58\% & 71.26\% \\
		FlowDetTrack(ours) & \textbf{39.64\%} & \textbf{40.90\%} & \textbf{38.72\%} & \textbf{72.04\%} \\
		\bottomrule
	\end{tabular}%
	\label{tab:test}%
\end{table}%

The robust in the presence of extrem motion with rotation is test on hard data set. The hard data set was obtained by adding random translations in the range of 0-0.5 m and random rotations around the Z-axis in the range of 0-20 degrees to each target on the KITTI MOT validation set. The results of car category and pedestrian category are show in table \ref{tab:hard}. When compared with the state-of-the-art 3D MOT method 3DMOT, the proposed method can still maintain high HOTA on the hard data sets.

\begin{table}[t]
	\centering
	\caption{Comparative Evaluation Results in Hard Datasets.}
	\begin{tabular}{ccccc}
		\toprule
		& HOTA$\uparrow$ & DetA$\uparrow$ & AssA$\uparrow$ & LocA$\uparrow$ \\
		\midrule
		\multicolumn{5}{c}{Car Category} \\
		\midrule
		AB3DMOT\cite{9341164} & 43.34\% & 58.07\% & 32.51\% & 88.08\% \\
		FlowDetTrack(ours) & 64.64\% & 67.00\% & 62.54\% & 88.59\% \\
		\midrule
		\multicolumn{5}{c}{Pedestrian Category} \\
		\midrule
		AB3DMOT\cite{9341164} & 15.14\% & 23.02\% & 10.08\% & 80.20\% \\
		FlowDetTrack(ours) & 32.21\% & 46.82\% & 22.28\% & 83.88\% \\
		\bottomrule
	\end{tabular}%
	\label{tab:hard}%
\end{table}%

%

\subsection{Qualitative Evaluation}
\begin{figure}[t]
	\begin{center}
		\includegraphics[width=0.45\textwidth]{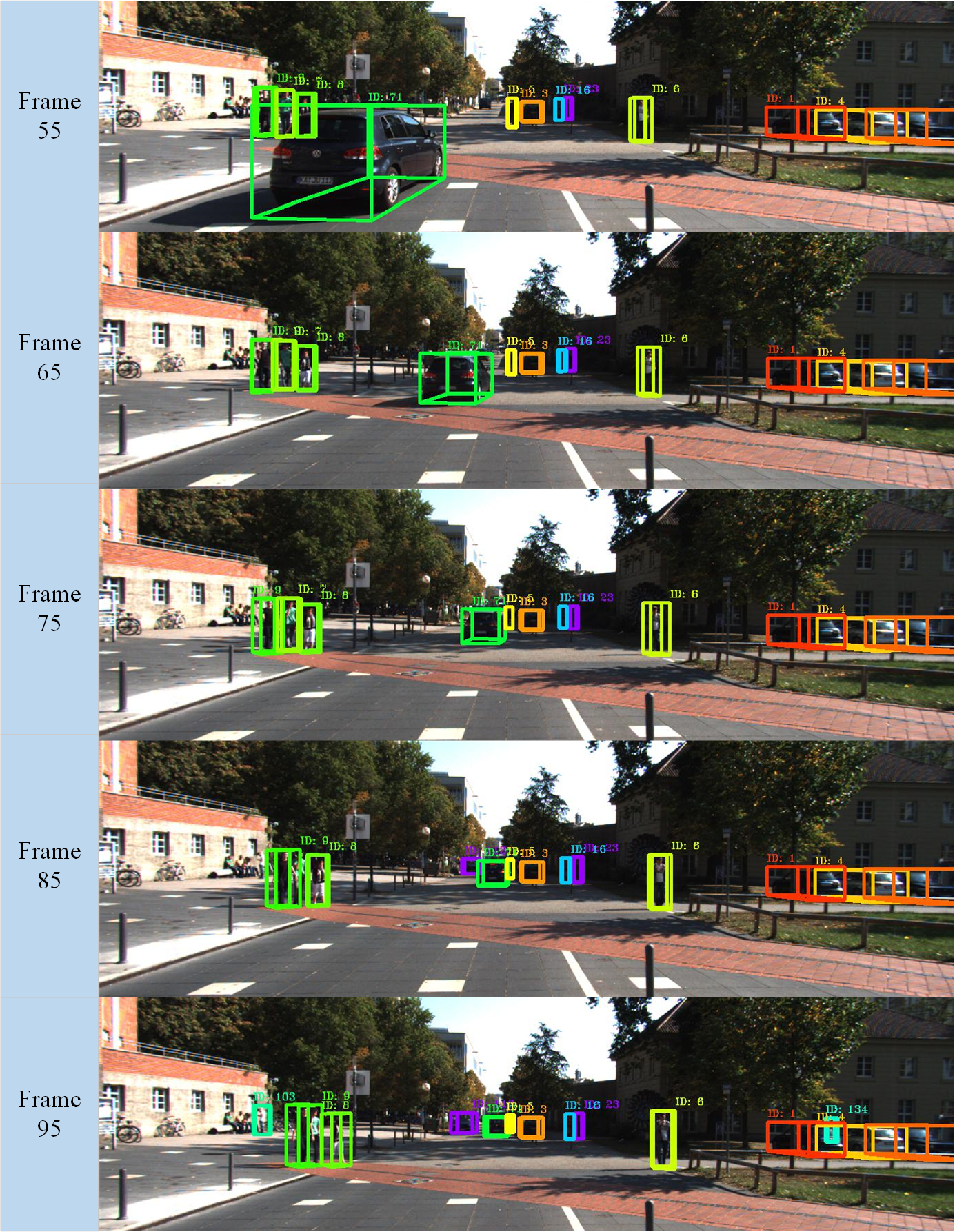}
	\end{center}
	\caption{The Visualization of MOT Results.}
	\label{MOT_vis}
\end{figure}

\begin{figure}[t]
	\begin{center}
		\includegraphics[width=0.5\textwidth]{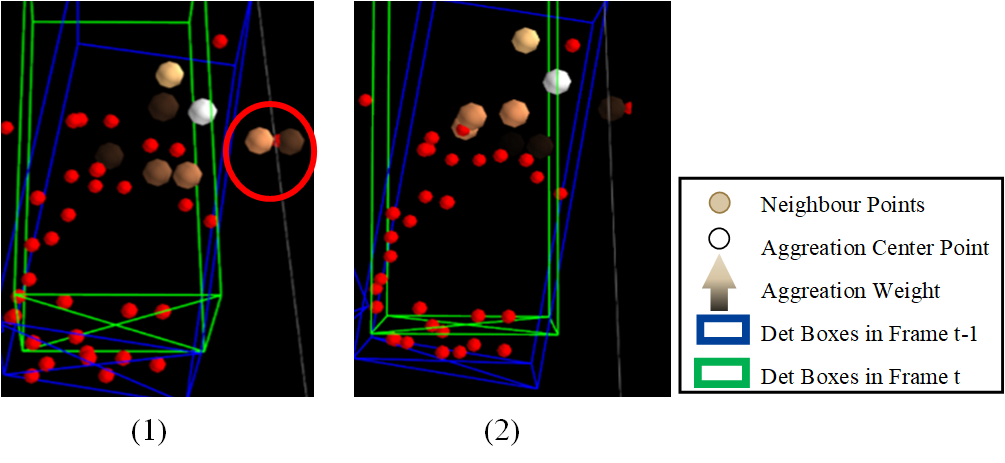}
	\end{center}
	\caption{The Visualization of Aggregation Weight Calculation. The subimage (1) is the result of weight calculation with distance in geometric space. The subimage (2) is the result of weight calculation with distance in both geometric space and detection result space.}
	\label{det_weight}
\end{figure}

\begin{figure}[t]
	\begin{center}
		\includegraphics[width=0.5\textwidth]{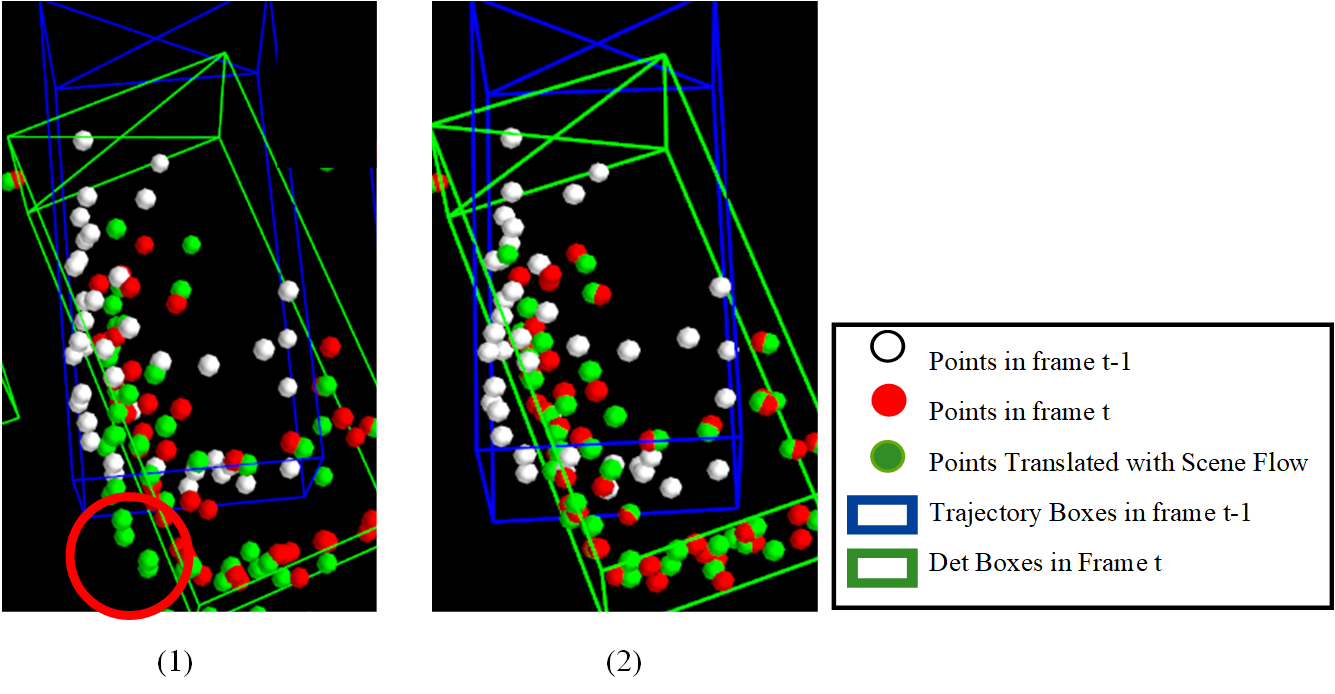}
	\end{center}
	\caption{The Visualization of Scene Flow Label Calculation. The subimage (1) is the result of box-translation-based method. The subimage (2) is the result of box-transformation-based method.}
	\label{sceneflow_label}
\end{figure}
Figure \ref{MOT_vis} is the visualization of the MOT results. Trajectory boxes are projected to the image. The box with the same tracking id is painted with the same color.

Figure \ref{det_weight} is the visualization result of the feature aggregation in the cost layer of scene flow module. When only the distance in geometric space is considered, those points that are close to the aggregation center but do not belong to the object will be given a large aggregation weight as indicated by the red circle in subimage (1). With the proposed detection-guidance feature aggregation methods, the aggregation weights of those background points will be small enough to be ignored. 

Figure \ref{sceneflow_label} shows the comparative visualization results of the scene flow calculation methods when the target has rotational motion. We translate the points in frame $t-1$ with the calculated scene flow ground truth label and the result is shown as green points. With box-translation-based method as in PointTrackNet\cite{9000527}, the green points as indicated by the red circle do not properly match the points in frame t. While, with the proposed box-transformation-based method, the green points can match the points in frame t nicely. This proves that the proposed box-transformation-based scene flow label calculation method can obtain a more accurate true value of scene flow in the case of target rotation.

\section{CONCLUSIONS}
We proposed a 3D MOT framework that optimizes the object detection and scene flow estimation simultaneously. A detection-guidance scene flow estimation is proposed to explore the role of object detection in promoting scene flow estimation. A box-transformation-based scene flow labels calculation method is proposed for more accurate scene flow labels in the case of targets' rotation.
The experimental results demonstrate that our network shows competitive results over the state-of-the-arts and robustness under extreme motion with rotation. 

\addtolength{\textheight}{-12cm}   


\bibliographystyle{IEEEtran}
\bibliography{IEEEabrv,ref}

\begin{thebibliography}{10}
\providecommand{\url}[1]{#1}
\csname url@samestyle\endcsname
\providecommand{\newblock}{\relax}
\providecommand{\bibinfo}[2]{#2}
\providecommand{\BIBentrySTDinterwordspacing}{\spaceskip=0pt\relax}
\providecommand{\BIBentryALTinterwordstretchfactor}{4}
\providecommand{\BIBentryALTinterwordspacing}{\spaceskip=\fontdimen2\font plus
\BIBentryALTinterwordstretchfactor\fontdimen3\font minus
  \fontdimen4\font\relax}
\providecommand{\BIBforeignlanguage}[2]{{%
\expandafter\ifx\csname l@#1\endcsname\relax
\typeout{** WARNING: IEEEtran.bst: No hyphenation pattern has been}%
\typeout{** loaded for the language `#1'. Using the pattern for}%
\typeout{** the default language instead.}%
\else
\language=\csname l@#1\endcsname
\fi
#2}}
\providecommand{\BIBdecl}{\relax}
\BIBdecl

\bibitem{chiu2020probabilistic}
H.-k. Chiu, A.~Prioletti, J.~Li, and J.~Bohg, ``Probabilistic 3d multi-object
  tracking for autonomous driving,'' \emph{arXiv preprint arXiv:2001.05673},
  2020.

\bibitem{9341164}
X.~Weng, J.~Wang, D.~Held, and K.~Kitani, ``3d multi-object tracking: A
  baseline and new evaluation metrics,'' in \emph{2020 IEEE/RSJ International
  Conference on Intelligent Robots and Systems (IROS)}, 2020, pp.
  10\,359--10\,366.

\bibitem{shenoi2020jrmot}
A.~Shenoi, M.~Patel, J.~Gwak, P.~Goebel, A.~Sadeghian, H.~Rezatofighi,
  R.~Mart{\'\i}n-Mart{\'\i}n, and S.~Savarese, ``Jrmot: A real-time 3d
  multi-object tracker and a new large-scale dataset,'' in \emph{2020 IEEE/RSJ
  International Conference on Intelligent Robots and Systems (IROS)}, 2020, pp.
  10\,335--10\,342.

\bibitem{zhai2020flowmot}
G.~Zhai, X.~Kong, J.~Cui, Y.~Liu, and Z.~Yang, ``Flowmot: 3d multi-object
  tracking by scene flow association,'' \emph{arXiv preprint arXiv:2012.07541},
  2020.

\bibitem{9157602}
X.~{Weng}, Y.~{Wang}, Y.~{Man}, and K.~M. {Kitani}, ``Gnn3dmot: Graph neural
  network for 3d multi-object tracking with 2d-3d multi-feature learning,'' in
  \emph{2020 IEEE/CVF Conference on Computer Vision and Pattern Recognition
  (CVPR)}, 2020, pp. 6498--6507.

\bibitem{7410673}
A.~Dosovitskiy, P.~Fischer, E.~Ilg, P.~Häusser, C.~Hazirbas, V.~Golkov,
  P.~v.~d. Smagt, D.~Cremers, and T.~Brox, ``Flownet: Learning optical flow
  with convolutional networks,'' in \emph{2015 IEEE International Conference on
  Computer Vision (ICCV)}, 2015, pp. 2758--2766.

\bibitem{8579029}
D.~Sun, X.~Yang, M.-Y. Liu, and J.~Kautz, ``Pwc-net: Cnns for optical flow
  using pyramid, warping, and cost volume,'' in \emph{2018 IEEE/CVF Conference
  on Computer Vision and Pattern Recognition}, 2018, pp. 8934--8943.

\bibitem{wu2019pointpwc}
W.~Wu, Z.~Wang, Z.~Li, W.~Liu, and L.~Fuxin, ``Pointpwc-net: A coarse-to-fine
  network for supervised and self-supervised scene flow estimation on 3d point
  clouds,'' \emph{arXiv preprint arXiv:1911.12408}, 2019.

\bibitem{8953876}
X.~Liu, C.~R. Qi, and L.~J. Guibas, ``Flownet3d: Learning scene flow in 3d
  point clouds,'' in \emph{2019 IEEE/CVF Conference on Computer Vision and
  Pattern Recognition (CVPR)}, 2019, pp. 529--537.

\bibitem{zhang2020multiple}
J.~Zhang, S.~Zhou, X.~Chang, F.~Wan, J.~Wang, Y.~Wu, and D.~Huang, ``Multiple
  object tracking by flowing and fusing,'' \emph{arXiv preprint
  arXiv:2001.11180}, 2020.

\bibitem{9000527}
S.~Wang, Y.~Sun, C.~Liu, and M.~Liu, ``Pointtracknet: An end-to-end network for
  3-d object detection and tracking from point clouds,'' \emph{IEEE Robotics
  and Automation Letters}, vol.~5, no.~2, pp. 3206--3212, 2020.

\bibitem{9157234}
S.~Shi, C.~Guo, L.~Jiang, Z.~Wang, J.~Shi, X.~Wang, and H.~Li, ``Pv-rcnn:
  Point-voxel feature set abstraction for 3d object detection,'' in \emph{2020
  IEEE/CVF Conference on Computer Vision and Pattern Recognition (CVPR)}, 2020,
  pp. 10\,526--10\,535.

\bibitem{8954200}
W.~{Wu}, Z.~{Qi}, and L.~{Fuxin}, ``Pointconv: Deep convolutional networks on
  3d point clouds,'' in \emph{2019 IEEE/CVF Conference on Computer Vision and
  Pattern Recognition (CVPR)}, 2019, pp. 9613--9622.

\bibitem{kuhn1955hungarian}
H.~Kuhn, ``The hungarian method for the assignment problem,'' \emph{Naval
  research logistics quarterly}, vol.~2, no. 1-2, pp. 83--97, 1955.

\bibitem{6248074}
A.~Geiger, P.~Lenz, and R.~Urtasun, ``Are we ready for autonomous driving? the
  kitti vision benchmark suite,'' in \emph{2012 IEEE Conference on Computer
  Vision and Pattern Recognition}, 2012, pp. 3354--3361.

\bibitem{luiten2021hota}
J.~Luiten, A.~Osep, P.~Dendorfer, P.~Torr, A.~Geiger, L.~Leal-Taix{\'e}, and
  B.~Leibe, ``Hota: A higher order metric for evaluating multi-object
  tracking,'' \emph{International journal of computer vision}, vol. 129, no.~2,
  pp. 548--578, 2021.

\bibitem{smith2019super}
L.~Smith and N.~Topin, ``Super-convergence: Very fast training of neural
  networks using large learning rates,'' in \emph{Artificial intelligence and
  machine learning for multi-domain operations applications}, vol. 11006.\hskip
  1em plus 0.5em minus 0.4em\relax International Society for Optics and
  Photonics, 2019, p. 1100612.

\bibitem{simon2019complexer}
M.~Simon, K.~Amende, A.~Kraus, J.~Honer, T.~Samann, H.~Kaulbersch, S.~Milz, and
  H.~Michael~Gross, ``Complexer-yolo: Real-time 3d object detection and
  tracking on semantic point clouds,'' in \emph{Proceedings of the IEEE/CVF
  Conference on Computer Vision and Pattern Recognition Workshops}, 2019, pp.
  0--0.

\end{thebibliography}


\end{document}